# BREAST CANCER DETECTION USING DEEP LEARNING

GAYATHRI G*
School of Electronics Engineering
Vellore Institute of Technology
Chennai, Tamil Nadu, India
gayathrig.2019@vitstudent.ac.in

PONNATHOTA SPANDANA*
School of Electronics Engineering
Vellore Institute of Technology
Chennai, Tamil Nadu, India
ponnathota.spandana2019@vitstudent.ac.in

BADRISH VASU*
School of Electronics Engineering
Vellore Institute of Technology
Chennai, Tamil Nadu, India
badrish.vasu2019@vitstudent.ac.in

SATHIYA NARAYANAN SEKAR
School of Electronics Engineering
Vellore Institute of Technology
Chennai, Tamil Nadu, India
sathiyanarayanan.s@vit.ac.in

*Abstract*— *Objective:* This paper proposes a deep learning model for breast cancer detection from reconstructed images of microwave imaging scan data and aims to improve the accuracy and efficiency of breast tumor detection, which could have a significant impact on breast cancer diagnosis and treatment. *Methods:* Our framework consists of different convolutional neural network (CNN) architectures for feature extraction and a region-based CNN for tumor detection. We use 7 different architectures: DenseNet201, ResNet50, InceptionV3, InceptionResNetV3, MobileNetV2, NASNetMobile and NASNetLarge and compare its performance to find the best architecture out of the seven. An experimental dataset of MRI-derived breast phantoms was used. *Results:* NASNetLarge is the best architecture which can be used for the CNN model with accuracy of 88.41% and loss of 27.82%. Given that the model's AUC is 0.786, it can be concluded that it is suitable for use in its present form, while it could be improved upon and trained on other datasets that are comparable. *Impact:* One of the main causes of death in women is breast cancer, and early identification is essential for enhancing the results for patients. Due to its non-invasiveness and capacity to produce high-resolution images, microwave imaging is a potential tool for breast cancer screening. The complexity of tumors makes it difficult to adequately detect them in microwave images. The results of this research show that deep learning has a lot of potential for breast cancer detection in microwave images.

## I. INTRODUCTION

Breast cancer detection using microwaves is a field that has been explored actively for a couple decades now [15]. The conventional screening methods such as mammography which performs poorly with dense tissues, exposes the patient's breasts to ionization radiation and is a painful process overall, which could be avoided by using microwave imaging [18]. Microwave imaging is an emerging medical imaging modality that has the potential to detect breast tumors at an early stage, very often used as a complementary approach to mammography [18], but its image reconstruction process is complex and prone to artifacts. In recent years, deep learning methods have shown remarkable success in various medical image analysis tasks, including tumor detection.

Microwave imaging uses low-power microwave systems which do not use ionizing radiation, making it safe for frequent usage. The contrast in dielectric properties of unhealthy/healthy breast tissues enables microwave imaging to be a low-cost, low-power, and non-invasive method to screen cancerous breast tissues [19].

Therefore, in this paper, we propose a deep learning framework for breast tumor detection from reconstructed images of microwave imaging scan data using different convolutional neural network architectures. By contributing to researching more on implementing microwave imaging effectively, eventually it can be used clinically and would usher medical screenings into a new era of efficiency and non-invasiveness.

In our work on the UM-BMID dataset, we used multiple models as our backbone model for feature extraction and classification. Some of these models include InceptionV3, ResNet50, MobileNetV2, InceptionResNetV2, DenseNet201 and NASNet. We chose DenseNet201 due to its ability to extract features from images at different scales and its effectiveness in capturing complex patterns and structures in medical images. InceptionV3 is a good option for medical image analysis because of its reputation for being able to capture various levels of abstraction in an image. Contrarily, ResNet50 was created to address the vanishing gradient issue that can arise in deep neural networks, enabling deeper networks with better performance. As a mobile and embedded device-optimized lightweight model, MobileNet is a good option for environments with limited resources.

While NASNet is a neural architecture search (NAS) model that uses reinforcement learning to automatically find neural network architectures that outperform human-designed models, InceptionResNet combines the strengths of both the Inception and ResNet architectures. A collection of 200 scans of MRI-derived breast phantoms provided the basis for the third generation of the University of Manitoba Breast Microwave Imaging Dataset (UM-BMID). The images were reconstructed using common beamforming methods like delay-and-sum, delay-multiply-and-sum, and far-field imaging algorithm using the scan data of third generation dataset. Afterwards, tumor detection wes performed on the reconstructed images.

## II. RELATED WORK

When compared to other forms of body tissue, breast tissue has unique electrical characteristics that can be used to discriminate between malignant and healthy tissue. As a result, there is a lot of knowledge in the scientific literature about the research of breast imaging techniques that are intended to find and diagnose breast cancer. Using the third generation of the UMID dataset, Tyson Reimer et al. [1] compared the images generated by a new method they proposed with those generated by the traditional delay-and-sum and delay-multiply-and-sum beamforming techniques. [2] assessed how well methods based on deep learning might identify cancers using breast microwave sensing. They compared it to a "dense neural network (DNN)" and "logistic regression classifier" of comparable size in order to detect the presence of malignant lesions in "experimental scans of MRI-derived phantoms". The sinogram data structure was used by the CNN to obtain a substantially greater level of diagnostic accuracy than random classification, outperforming the other classifiers in the detection of cancer. Priyam Patel et al. [3] analyzed a dataset's accessibility and used a variety of machine learning classification algorithms to find tumors in the UM-BMID dataset. They then compared their results with previously published findings. [4] utilized "a rejection model based on Support Vector Machine (SVM)" to reduce the "false positive (FP)" rate of divided mammogram images obtained from the "Chan-Vese method". "The Marker Controller Watershed (MCWS) algorithm" was used to initialize the segmentation process. The experimental results demonstrated that incorporating the "SVM rejection model" was effective in decreasing the FP rate as contrasted to the results obtained without it. [5] introduces "a composite end-to-end framework" that integrates "convolutional neural network (CNN) and long-short-term memory (LSTM)" to perform two tasks which are detecting and quadrant locating breast tumors. This framework does not require complex microwave imaging processing. Shubham Sharma et al. [6] conducted a comparison of commonly used "machine learning algorithms and techniques" for predicting breast tumor, namely "Random Forest, kNN (k-Nearest-Neighbor), and Naïve Bayes". They utilized the "Wisconsin Diagnosis Breast Cancer data" set as a training set to assess the performance of these techniques based on key parameters like accuracy and precision. [16] To make the "process of tumor diagnosis, localization, and characterisation from scatter parameters measures" and metadata information easier, Al Khatib, Salwa K., et al. proposes a deep learning architecture made up of "deep neural networks using convolutional layers". The created deep learning framework performs better than existing methods in literature in terms of "tumor localization, characterisation, and detection accuracy". For breast microwave imaging, this paper[17] provides "a novel iterative delay-and-sum oriented reconstruction approach". An image estimate is iteratively updated by the "iterative DAS algorithm" using a forward radar model. Additionally, a delay-multiply-and-sum beamformer based "itDAS reconstruction method variant" was put into practice.

## III. PROPOSED SYSTEM

In the UM-BMID dataset breast phantom scans, S11 and S21 parameters are collected using antennas to measure the electromagnetic waves that are scattered by the breast phantom. An antenna's scattering behavior is described using the S11 and S21 parameters. In order to create images from the S11 and S21 scans, the acquired data must first be processed to get rid of any noise and artifacts that might have been added during the measuring procedure. For image reconstruction, the traditional methods of delay-and-sum, delay-multiply-and-sum are utilized. Reconstructed image is created by merging the data obtained from several ultrasound transducer elements in both DAS and DMAS algorithms. The transducer array produces a succession of ultrasound waves that travel through the tissue and are reflected, scattered, and absorbed by the different tissue structures. The transducer elements track the returning waves, and from their amplitude and time delay, it is possible to recreate an image of the tissue structure.

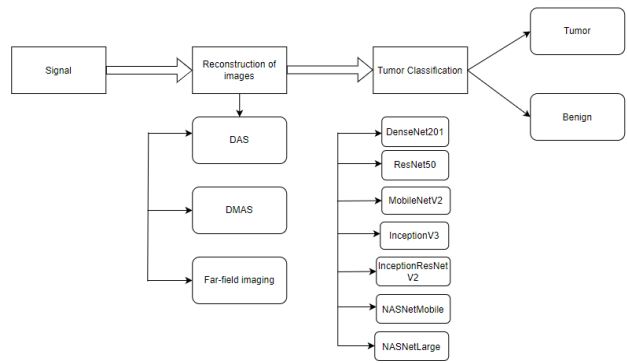

Fig.1. Proposed framework of Breast cancer detection

For generating a three-dimensional image, the delay-and-sum (DAS) beamforming technique is utilized in medical imaging. The procedure entails first processing the signals that the transmitting, receiving antennas have sent and received. Following that, based on the location of the antenna, the location of the focal point, and a prediction of the wave propagation speed, the proper time-delays are determined for each received signal. Spatial beamforming is produced as a result of the focal point's movement within the breast during the focusing procedure. Each site performs a coherent summation and integration of all time-shifted replies as shown in Fig. 2. The windowed signal is integrated, with the width of the integration window determined by the system bandwidth. A three-dimensional representation of the scattered energy is created by adding up and integrating the time-shifted replies. With the use of this image, anomalies or diseases of the breast can be found.

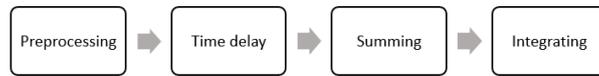

Fig.2. Block diagram of Delay-and-Sum (DAS)

Delay-Multiply-and-Sum (DMAS) is a reconstruction technique used in medical imaging to produce high-quality images with reduced artifacts. When using this method, the signals that were received are first multiplied by a weight function that takes into account the frequency-dependent attenuation properties of the medium the signal has passed through. The signals are then appropriately time-delayed in accordance with the placement of the focal point, the transmitting and receiving antennas. The delayed and weighted signals are then multiplied by one another, producing a multiplication of each signal by both itself and the other signals.

This multiplication step is what sets DMAS apart from DAS, where signals are just added together as mentioned in Fig.3. The picture artifacts and reverberations are less noticeable due to the multiplication. In order to create an image of the breast, the results of the multiplication step are finally added across all of the time delays and focal points. Medical experts can use this technology to generate high-resolution images with fewer artifacts, which makes it an important tool for the detection and management of breast illnesses.

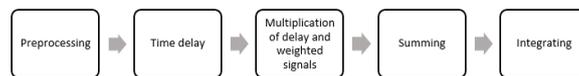

Fig.3. Block diagram of Delay-Multiply-and-Sum (DMAS)

Additionally, we also used a far-field imaging algorithm to generate reconstructed images from scan data. [1]Radar-based image reconstruction methods like DAS and DMAS only require one mathematical step to produce an image. These techniques are rapid, but they are unable to handle some complicated elements of the radar signal, such as nonlinear effects. The issue with these methods is that they presume that the radar signal moves directly from the antenna to the item being observed and back again. Additionally, they believe that the signal won't be lost or altered along the way. However, these presumptions might not hold true in real-world situations. The signal might be reduced by distance or only partially transmitted at interfaces, or it might be impacted by various propagation rates inside the imaging domain. Distances between the antenna and each scattering point as well as between the antenna and each scattering point are computed for the image reconstruction process.

DenseNet Architecture:
DenseNet201 is a convolutional neural network (CNN) architecture that was proposed by Huang et al. in 2017. It is one of the more powerful members of the DenseNet family of architectures, which are characterized by their dense connectivity patterns between layers as shown in Fig. 4. Dense connectivity refers to a feed-forward connection between every layer and every other layer. DenseNet201 consists of multiple dense blocks, where each dense block contains a set of convolutional layers that are connected to each other. The output of each dense block is then passed to a transition layer, which reduces the spatial dimensions of the feature maps by applying a pooling operation and a convolutional layer. The final dense block is followed by a global average pooling layer, which averages the feature maps over their spatial dimensions and produces a one-dimensional vector that is fed to a fully connected softmax layer for classification.

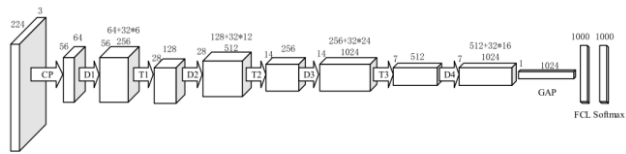

Fig.4. Densenet 201 architecture[8]

ResNet50 Architecture:

A deep neural network architecture called ResNet50 was created for image identification tasks. It was created by Microsoft Research Asia and took first place in the COCO and ImageNet image recognition challenges in 2015. The introduction of residual blocks, which allow the model to develop residual mappings between the input and output feature maps, is the main novelty of the ResNet design. Convolutional, pooling, and fully linked classification layers are among the 50 layers of the

ResNet50 design as shown in Fig. 5. To extract local and global characteristics from the input image, it combines 3x3 and 1x1 convolutional filters. The ResNet50 architecture's residual blocks are made up of convolutional layers that are followed by shortcut connections that omit one or more layers.

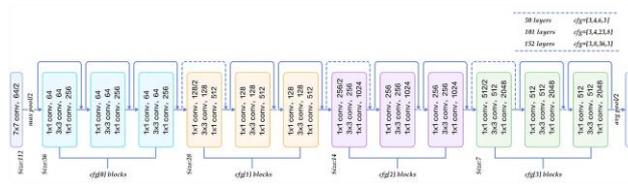

Fig.5. ResNet50 Architecture [10]

MobileNetV2 Architecture:

MobileNetV2, a deep neural network architecture, is intended for mobile and embedded vision applications.. It was created by the Mobile Vision team at Google and is an expansion of the previous MobileNet architecture, originally created for low-power and minimal latency devices. In order to achieve depthwise separability, the MobileNetV2 design combines linear bottlenecks with convolutions. Additionally, there are linear bottlenecks in the MobileNetV2 design, which are 1x1 convolutional layers that limit the amount of channels in the feature maps as shown in Fig. 6. The linear bottlenecks increase the network's non-linearity, which lowers the model's computing cost and boosts accuracy.

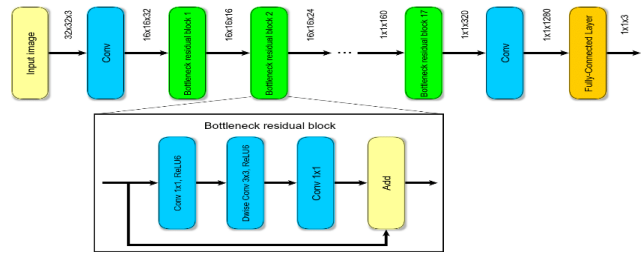

Fig.6. MobileNetV2 Architecture[11]

InceptionV3 Architecture:

A deep neural network architecture called InceptionV3 was created for image identification tasks. It is a development of the initial Inception architecture that was unveiled in 2014 and was created by Google's research team. Convolutional layers, pooling layers, and inception modules all play a part in the foundation of the InceptionV3 architecture. A group of parallel convolutional layers of various widths, followed by a layer that pools them and concatenation of their outputs, make up the inception module, a building block as shown in Fig. 7. The model can now recognize both local and global elements in the input image thanks to this approach.

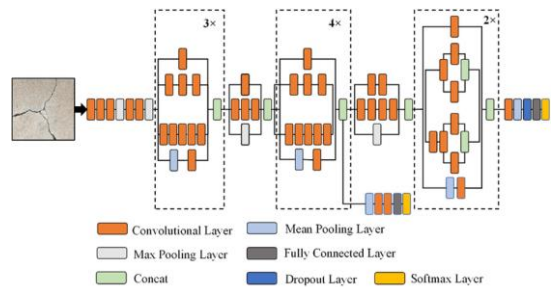

Fig.7. InceptionV3 Architecture[12]

InceptionResNetV2 Architecture:

A deep neural network architecture called InceptionResNetV2 was created for image identification challenges. It is an extension to the Inception architecture containing residual connections that was created by Google's research team. It produces a model that is more accurate and efficient by combining the advantages of the Inception and ResNet designs. A number of inception modules are joined via residual connections to form the InceptionResNetV2 architecture. Multiple convolutional layers with various filter sizes make up each inception module, which then proceed by batch normalization, ReLU activation, and max pooling as shown in Fig. 8.

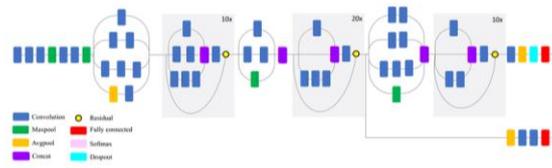

Fig.8. InceptionResNetV2 Architecture[13]

NASNetMobile Architecture:

A deep neural network architecture called NASNetMobile was created for use in embedded and mobile vision applications. In order to automatically find the best architecture for image recognition tasks, Google's research team created it utilizing neural architecture search (NAS) methods. A sequence of stacked cells make up the NASNetMobile architecture, and each cell combines normal and reduction processes. While the reduction operation applies a mixture of 3x3 max pooling and 1x1 convolution, the normal operation employs a combination of 3x3 and 1x1 convolutions.

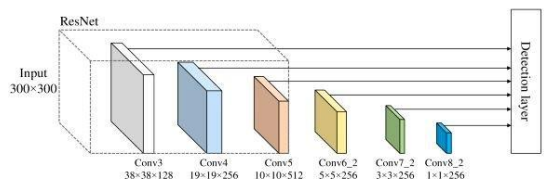

Fig.9. NASNetMobile Architecture[14]

NASNetLarge Architecture

The Google Brain AutoML team used neural architecture search techniques to create the deep neural network architecture known as NASNetLarge, which is intended for image recognition applications. On huge datasets, the architecture, which is built on a cell-based search space, is designed for excellent accuracy and efficiency. The NASNetLarge design is made up of a stack of repeating cells, each of which has a set of operations that may be mixed and matched in many ways to create new architectures. Convolutional layers, pooling layers, and other non-linear procedures like batch normalization and ReLU are among the operations.

Over 5.3 million parameters and 88 convolutional layers comprise the NASNetLarge architecture. For classification, it combines fully connected layers and global average pooling. A skip connection in the design enables the model to pick up on both local and global elements in the input image. The NASNetLarge architecture uses regular cells and reduction cells, which is one of its distinctive features. The reduction cells are employed to downsample the feature maps and collect global image features, while the normal cells are used to acquire local image features. This enhances the model's effectiveness and accuracy.

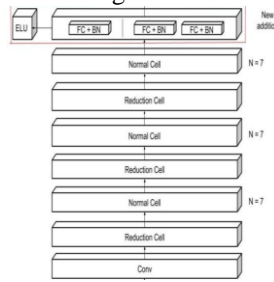

Fig.10. NASNetLarge architecture. FC is fully connected layers and BN is Batch Normalization[9]

All the architectures used are trained on the ImageNet dataset to enable transfer learning. Transfer Learning enables utilization of a pretrained network, ImageNet in our case, to perform new classification/prediction tasks. Transfer learning offers an increase in performance and saves time during training. [15]

## IV. DATASET

A dataset called UM-BMID is made up of three generations of data, each of which contains details from scans of breast phantoms with various characteristics. The first generation dataset includes information from scans of 13 distinct phantoms made from three different types of polycarbonate-printed adipose shells. Data from scans of 66 distinct phantoms made from 9 different adipose shells printed with polycarbonate copolyester are incorporated in the "second generation dataset". "The distributions of BI-RADS density class, tumor size, and breast size" are essentially uniform between generations. The second generation dataset has unique scans for each phantom with respect to those factors as well as "adipose shell position, fibroglandular shell position, and rotation". The first generation dataset has unique scans for each phantom with respect to "breast phantom I.D., tumor size, and tumor position".[7]

The dataset's first and second generations utility is limited by issues with "target location uncertainty and reference scan acquisition". The third generation of UM-BMID was created using 3D-printed phantoms scanned with a better Breast Microwave Sensing(BMS) system that runs over a wider frequency range and has a greater frequency bandwidth in order to get around these restrictions. "The third-generation dataset comprises of scans of 20 different geometries of healthy phantom tissue", each of which was scanned five times with tumors of varying diameters (10mm, 15mm, 20mm, 25mm, and 30mm in diameter) positioned in the same location in each of the five scans. "The antenna was rotated to 72 points evenly spaced along a circular trajectory covering 355 degrees" as part of the scan routine, which required obtaining measurements at 1001 frequencies over the 1-9 GHz spectrum. To simplify image-based tumor identification analysis in experimental BMS, the third generation UM-BMID was developed.[1]

|  | Gen - 1 | Gen - 2 | Gen - 3 |
| --- | --- | --- | --- |
| Total number of scans | 249 | 1008 | 200 |
| Measured Parameters | $S_{11}$ | $S_{11}, S_{21}$ | $S_{11}, S_{21}$ |
| Scan Frequencies | 1-8 GHz | 1-8 GHz | 1-9 GHz |
| Tumour diameters | 10, 20, 30 | 10, 20, 30 | 10, 15, 20, 25, 30 |

Table.1. Description of UMID Dataset[1]

## V. RESULTS & INFERENCES

The UM-BMID dataset is broken into two parts, training (820 images), validation (80 images). The dataset is put through a data augmentation function consisting of random rotations and zoom factors. Each image in the dataset is resized to 224x224 pixels. And these pixel values are then converted into an array and given to the model as input. Pre-trained ImageNet weights are used to enable Transfer Learning. Epochs varying from 10 to 15 were chosen. Adaptive learning rate from 4e-5 to 1e-7 was selected. After running the proposed architectures over the dataset, outcomes are as follows:

Validation Graphs:

i) DenseNet201

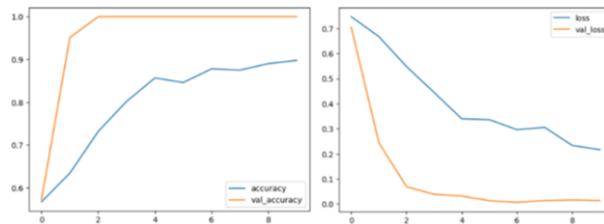

Fig.11. DenseNet201 accuracy vs validation accuracy
Fig.12. DenseNet201 loss vs validation loss

ii) ResNet50

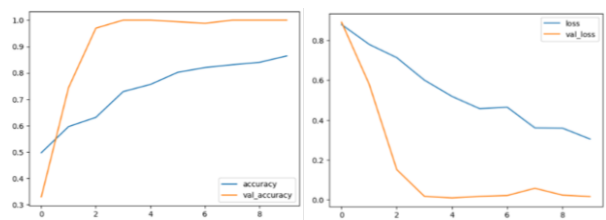

Fig.13. ResNet50 accuracy vs validation accuracy
Fig.14. ResNet50 loss vs validation loss

iii) MobileNetV2

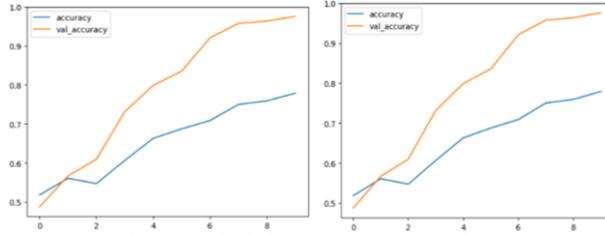

Fig.15. MobileNetV2 accuracy vs validation accuracy
Fig.16. MobileNetV2 loss vs validation loss

iv) InceptionV3

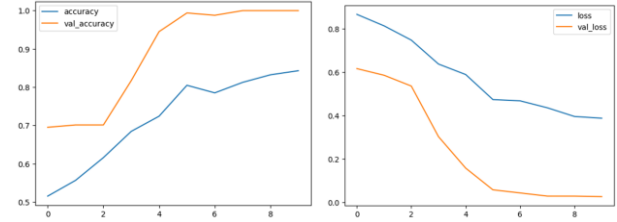

Fig.17. InceptionV3 accuracy vs validation accuracy
Fig.18. InceptionV3 loss vs validation loss

v) InceptionResNetV2

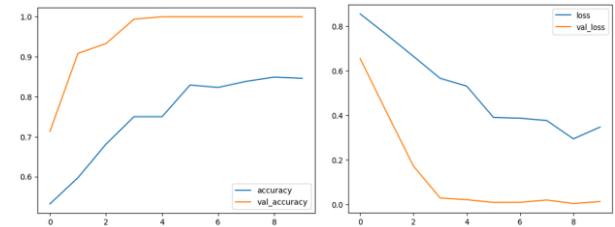

Fig.19. InceptionResNetV2 accuracy vs validation accuracy
Fig.20. InceptionResNetV2 loss vs validation loss

vi) NASNetMobile

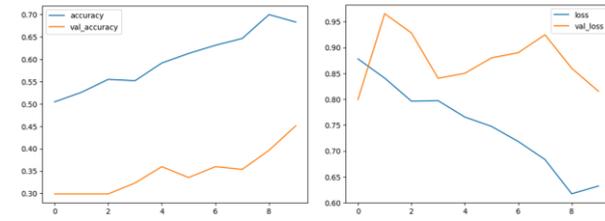

Fig.21. NASNetMobile accuracy vs validation accuracy
Fig.22. NASNetMobile loss vs validation loss

vii) NASNetLarge

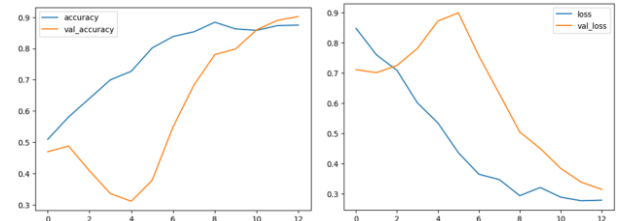

Fig.23. NASNetLarge accuracy vs validation accuracy

Fig.24. NASNetLarge loss vs validation loss

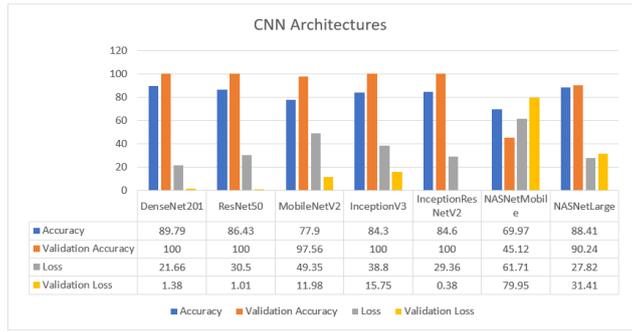

Table.2. Table comparing the performances of different architectures

Majority of the architectures used (DenseNet201, ResNet50, InceptionV3, InceptionResNetV2, NASNetMobile) cause the model to overfit the training data. A major factor contributing to this problem is the small size of the dataset (900 images) and the complexity of the dataset being too high which impedes the model's ability to learn effectively. However, MobileNetV2 showed the lowest overfitting after NASNetLarge and could potentially be tuned better to the dataset. Currently, only the NASNetLarge architecture has the lowest amount of overfitting and a decent accuracy (88.41%)

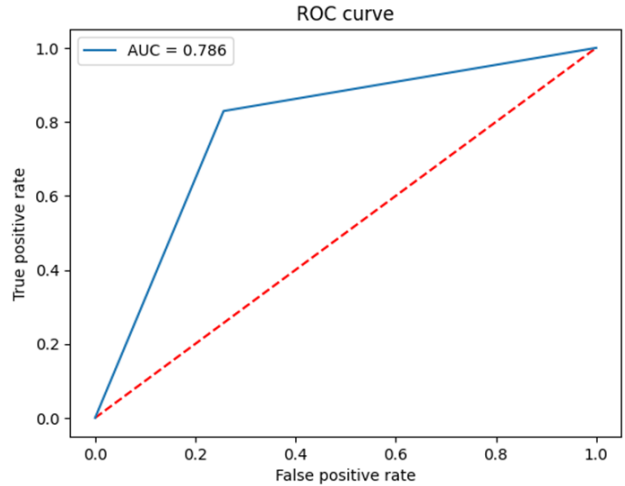

Fig,25. NASNetLarge AUC score

## VI. CONCLUSION AND FUTURE WORK

From the graphs and the chart, it can be seen that architecture DenseNet201 cannot be used despite having the highest accuracy due to the problem of overfitting. Therefore, it can be concluded that NASNetLarge is the best architecture which can be used for the CNN model with Accuracy = 88.41%, Loss = 27.82%. Since the model has an AUC of 0.786, it can be inferred that the model is acceptable enough to be used in its current state, although it could be fine tuned further and can be used to train on other similar datasets.

We intend to further this work by exploring various techniques to improve the accuracy and robustness of the models. Model ensemble is one possible strategy, which involves training multiple models and combining the results to make a final prediction. It has been demonstrated that ensemble methods can enhance the performance of machine learning models in a variety of applications, including medical image analysis.

Explainability and interpretability will additionally be a focus of future research on the UM-BMID dataset. It becomes more crucial to provide insights into the model's decision-making process as the algorithms become more complex. The regions of the image that are most essential to the diagnosis can be highlighted by using techniques like saliency maps, attention mechanisms, and visualization techniques.

Additionally, we want to incorporate electronic health records (EHRs). The models must be tested in clinical trials and validated on bigger, more varied datasets in order to be clinically useful. Integration with EHRs can help to improve the speed and accuracy of healthcare delivery as well as the diagnostic process. We aim to develop models that are user-friendly and simple for medical professionals to understand in order to be successfully incorporated into clinical workflow.